\documentclass[aps,12pt,a4paper]{revtex4}
\usepackage{epsfig}
\usepackage{graphicx}
\usepackage{amsmath,amssymb,color}
\usepackage[english]{babel}

\parskip=\medskipamount

\newcommand{\eq}[1]{(\ref{#1})}
\newcommand{\fig}[1]{Fig.~\ref{#1}}

\newcommand{\be}{\begin{equation}}
\newcommand{\ee}{\end{equation}}

\newcommand\disp{\displaystyle}

\begin{document}

\title{Analysis of English free association network reveals mechanisms of efficient solution of  Remote Association Tests}

\author{O.V. Valba$^1$, A.S. Gorsky$^{2,3}$, S.K. Nechaev$^{4,5}$, M.V. Tamm$^{1,6}$}

\affiliation{ $^1$ Department of Applied Mathematics, MIEM, National Research University Higher School of Economics, 123458, Moscow, Russia, \\ $^2$ Kharkevich Institute for Information Transmission Problems RAS, \\ $^3$Moscow Institute of Physics and Technology, Dolgoprudny 141700, Russia \\ $^4$ Interdisciplinary Scientific Center Poncelet (CNRS UMI 2615), 119002 Moscow, Russia \\ $^5$ P.N. Lebedev Physical Institute RAS, 119991 Moscow, Russia \\ $^6$ Faculty of Physics, Lomonosov Moscow State University, 119992 Moscow, Russia}

\begin{abstract}

We study correlations between the structure and properties of a free association network of the English language, and solutions of psycholinguistic Remote Association Tests (RATs). We show that average hardness of individual RATs is largely determined by relative positions of test words (stimuli and response) on the free association network. We argue that the solution of RATs can be interpreted as a first passage search problem on a network whose vertices are words and links are associations between words. We propose different heuristic search algorithms and demonstrate that in "easily-solving" RATs (those that are solved in 15 seconds by more than 64\% subjects) the solution is governed by "strong" network links (i.e. strong associations) directly connecting stimuli and response, and thus the efficient strategy consist in activating such strong links. In turn, the most efficient mechanism of solving medium and hard RATs consists of preferentially following sequence of "moderately weak" associations.

\end{abstract}

\maketitle

\section{Introduction}

Representation of a large number of interacting agents by a network is one of the most powerful ways of efficient treatment of various types of data in biological, technological, and social systems \cite{newman_book,barabasi_book}, as well as in cognitive processes.  A network is a set of nodes (the elementary indivisible units of a distributed system) and binary relations (links) between them. There is a plenty of ways to build networks in the cognitive science, with various setups relevant for a problem-dependent specific conditions. Historically, semantic networks were used to represent a "knowledge" by establishing directed or undirected semantic relations (graph links) between the "concepts" (graph nodes) \cite{grainger2016}. Such networks are useful to study the "intermediate" (or "mesoscopic") scale of organization in the human cognition. However, in attempts to model cognitive processes, it has been realized that the "microscale" network organization, i.e the structure of the detailed concept-to-concept connections, is very important.

Advances in graph-theoretic methods of studying cognitive functions are inextricably linked with pioneering works \cite{mendes2001, steyvers2005, vitevitch2008}. Since then, the number of investigations in the field has grown rapidly, and in particular a lot of attention has been paid to the study of large-scale semantic networks. In such networks, words (e.g., nouns) are nodes connected by links indicating semantic relations between them. There is a variety of characteristics of semantic proximity: one can connect the nearest neighboring words in sentences (so-called syntactic networks), or one can connect words according to standard linguistic relations between them (synonymy, hyper- or hyponymy, etc.). Finally, one can assemble networks of words based on various psycholinguistic experimental data.

Large-scale semantic networks possess a specific pattern of connectivity, presumably imposed by the growth processes by which these networks are formed. Typically, such networks demonstrate a power-law structure in the distribution of links across the nodes, such that most network nodes have a low vertex degree, however there are some nodes with a very high degree -- they play the role of hubs.

Important and fast-growing area in the field of linguistic networks is related to the so-called "word embedding" \cite{Mikolov2013}. "Word embedding" is a set of language-modeling techniques based on mapping of words to vectors of numbers, usually in a multidimensional Eucledian space. The semantic similarity of two words is defined as a scalar product of the corresponding vectors. Such a procedure results in construction of a complete weighted network of words (each pair of words is connected by a weighted edge, whose weight is the semantic proximity) and the majority of edges have very small weights. Removing all links with weights less than a preset threshold, results in a network with nontrivial topological properties. It might be very productive to generalize the "word embedding" ideology to non-Eucledian spaces, in particular to spaces of constant negative curvature which are natural target spaces for scale-free networks \cite{krioukov10, krioukov12}. The detailed discussion of these questions goes beyond the scope of the current work and will be published separately \cite{kasyanov_inprep}. To complete this short overview of various theoretic approaches, let us mention that recently several attempts have been made to treat semantic networks as multiplexes (i.e. multilayer networks). Such approaches seem to give deeper insight into the formation of mental lexicon \cite{stella2018} and early word acquisition \cite{stella2019}.

One particularly interesting class of semantic networks is a network of free associations \cite{kiss1973,nelson2004,dedeyne2019,russian_tesaurus,russian_kartaslov}. That is a class of networks obtained in the following real experiment. Participants ("test subjects") receive words ("stimuli") and asked to return, for each stimulus, the first word coming to their mind in response. The responses of many test subjects are aggregated and a directed network of weighted links between the words (stimuli and responses) reflecting the frequencies of answers is constructed. The study of these networks has a long history \cite{kiss1973,nelson2004}. In what follows, we use a free association network obtained in the "English Small World of Words" project (SWOW-EN) \cite{dedeyne2019}. The online data collection procedure allowed the authors of \cite{dedeyne2019} to aggregate data for more than 12~000 stimuli words. The data was collected in 2011-18 and consists of responses of more than 90~000 test subjects. As a result, the network contains many weak (i.e. rare) associations, which were not registered in earlier experiments.

In our work we use free associations network and propose some heuristic mechanisms of solving the so-called Remote Associates Test (RAT). The RAT has been invented by S. Mednick in 1962 \cite{mednik1962} and later has been repeatedly used in cognitive neuroscience and psychology \cite{kounious2014,helie2010,jung2004} to study insight, problem solving and creative thinking. The RAT test subjects are given sets of three stimuli words (e.g. "surprise", "line", "birthday") and are requested to find a fourth "return" word, which is simultaneously associatively related to all three stimuli (in our example it is the word "party").

Possible mechanisms of RAT solving have been extensively discussed in the literature. In \cite{smith2013} authors analyzed sequences of guesses, which came to mind during RAT solving. They measured the similarity between guesses, stimuli, and responses using the "Latent Semantic Analysis" \cite{smith2013} and concluded that there are two systematic strategies of solving multiply constrained problems. In the first strategy generation of guesses is primarily based on just one of three stimuli, while in the second strategy it is implied that the test subject is making new guesses based partially on his/her previous guesses. In \cite{bourgin2014} the Metropolis-Hastings search model has been used, which involved the transition probabilities based on geodesic (shortest) distances along the network from the stimuli to the response. The authors underline the importance of association strength between words in the process of RAT solving. The work \cite{olteteanu2015} is devoted to the design, implementation and analysis of a computational solver, which can answer RAT queries in a cognitively inspired manner. In \cite{olteteanu2015} authors developed an artificial cognitive system based on an unified framework of knowledge organization and processing. They took into account associative links between the concepts in the knowledge base and the frequency of their appearance. In a later work \cite{olteteanu2017}, it has been shown that the association strength and the number of associations both have important separated effects on performance of RAT solving. Finally, the spiking neural network model is proposed in \cite{kajic2017}. There, RAT solving is simulated as a superposition of two cognitive processes: the first one generates potential responses, while the second one filters the responses.

In our study we address two main questions. First, we study connection between the average hardness of a particular RAT and the position of stimuli and response on the free association network. We show that the RAT hardness can be predicted reasonably well by examining the network structure. Second, we discuss possible heuristic cognitive search algorithms of solving RATs, and study ways of their optimization.

The paper is organized as follows. In Section II we provide a brief characteristic of used datasets: the structural properties of the free association network, and the quantitative definition of RAT hardness. In Section III we study correlations between the RAT hardness and the relative position of stimuli and response on the free association network. We show that the RAT hardness significantly correlates with the aggregated weight of directed bonds stimuli $\to$ response, as well as with the aggregated weight of multi-step chains of associations. On the other hand, there is essentially no correlation between the RAT hardness and the weights of reverse (response $\to$ stimuli) bonds. We argue that such an asymmetry can be interpreted as a sign that solving a RAT is a first-passage problem: the correct response is easy to identify as soon as one finds it along a directed path on the network. In Section IV we study various ways of enhancing the probability of a fast solution of a RAT. We argue that the search strategies with resetting seem to be preferred for both nearest-neighbour search, and the search by infinitely long chains of associations. We discuss in details the role of weak associations in the search. In particular, we show that the best strategy for solving easy RATs implies removing all weak associations, and following only the strong ones. In turn, solving medium and especially hard RATs in the same way is often impossible. Instead, the probability to find a solution of hard RATs gets enhanced when the search runs preferentially along moderately weak bonds (associations). In Discussion we summarize the obtained results and propose possible direction of further investigations.

\section{Data analysis}

We use the free association network described in \cite{dedeyne2019} and known as "English Small World of Words project" (SWOW-EN). It is a weighted directed network with $N=12~217$ stimuli words. The brief summary of the network topological characteristics is presented in Table \ref{tab1}.

\begin{table}[ht]
\caption{The main topological properties of SWOW-EN.}
\begin{center}
\begin{tabular}{ll}
\hline \hline
Nodes & \hspace{1cm} 12~217 \\ \hline
Mean degree & \hspace{1cm} 31.7 \\ \hline
Diameter & \hspace{1cm} 8 \\ \hline
Transitivity & \hspace{1cm} 0.08 \\ \hline
Percolation threshold & \hspace{1cm} 0.08 \\ \hline\hline
\end{tabular}
\end{center}
\label{tab1}
\end{table}

\begin{figure}[ht]
\centerline{\includegraphics[width=16cm]{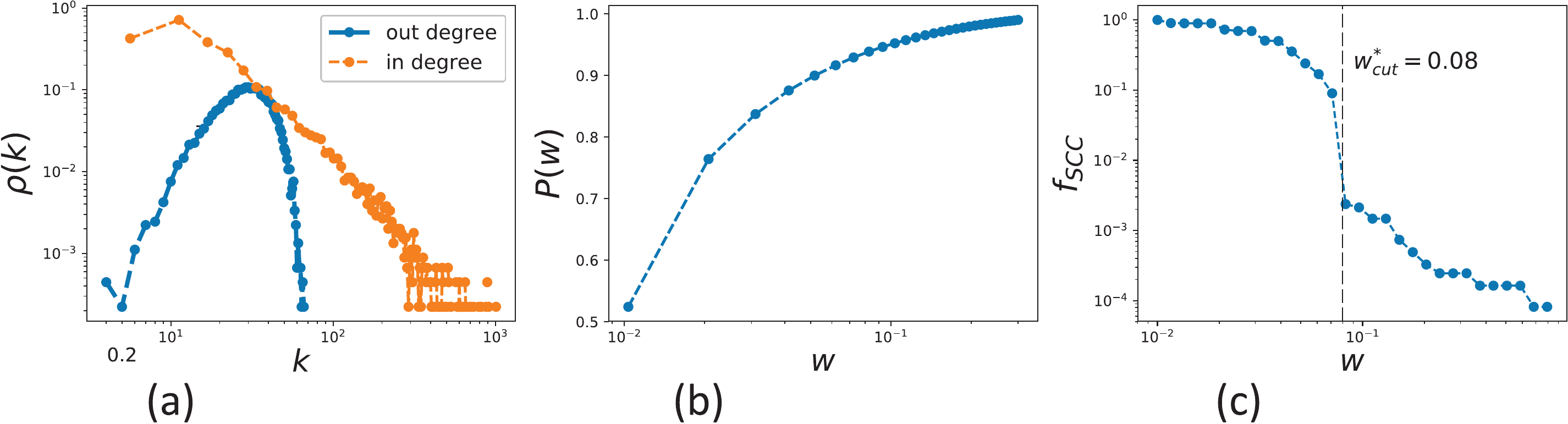}}
\caption{(a) Distributions of in- (blue) and out- (orange) vertex degrees of SWOW-EN; (b) Cumulative distribution of link weights of SWOW-EN; (c) Fraction of nodes in strongly connected component of SWOW-EN in dependence on the link weight cutoff (see \cite{dedeyne2019}).}
\label{fig01}
\end{figure}

In \fig{fig01}a we show the distribution of in- and out-degrees of the network. The out-degree distribution (blue) is Poissonian, its average is controlled by the experimental setup: the bigger the number of test subjects per stimulus word, the larger the average degree. In turn, the in-degree distribution $\rho(k)$ (orange), where $k$ is the vertex degree of incoming bonds, has a power-law tail, i.e. $\rho(k) \sim k^{-\gamma}$ with $\gamma \approx 3$. Interestingly, such a shape of the in-degree distribution seems to be quite universal: similar values of $\gamma$ are observed for networks in other experiments with English networks \cite{kiss1973,nelson2004}, and also for the Russian free association networks \cite{russian_tesaurus, russian_kartaslov}. In \fig{fig01}b we demonstrate the cumulative distribution $P(w)$ of weights $w$ of links of SWOW-EN (all weights lie within the interval $[0.01,1.00]$) and in \fig{fig01}c we depict the size of largest strongly connected component of SWOW-EN as a function of the link weight cutoff, $w_{cut}$. Note that the size of strongly connected component of SWOW-EN collapses at the link cutoff about $w_{cut}^*=0.08$, which corresponds to removal of 95\% of network edges. The SWOW-EN above $w_{cut}^*$ is not percolating anymore, and splits in disjoint components.

We use data for the hardness of RATs solution provided in \cite{bowden2003}. We restrict ourselves to 138  problems (combinations of three stimuli and response) out of 144 studied in \cite{bowden2003}, for which all four words are present in SWOW-EN network and have non-zero out-degree. The hardness of a RAT is quantitatively characterized by the value $H$ ($0\le H\le 1$), measuring the fraction of test subjects who correctly solved it in 15 seconds (for each of 138 problems under consideration). Additionally, we divide problems into three broad categories: ``easy", ``medium", and ``hard". The problem is easy if it has been solved in 15 seconds by more than 64\% test subjects ($0.64 \le H \le 1$), medium if it was solved by $32\% \div 64\%$ test subjects ($0.32 \le H \le 64$), and hard otherwise ($0 \le H \le 0.32$). There are 15 easy, 38 medium, and 85 hard problems. For completeness, we have reproduced in Appendix A used dataset of 138 problems (out of 144 presented in \cite{bowden2003}).

\section{Correlation between average RAT hardness and weights of edges in a free
association network}

The strength of an association between two given words (vertices) in a free association network, $G$, is described quantitatively by the weight of the corresponding directed link. The whole set of weights is encoded in the weighted adjacency matrix, $W(G)$. The element, $w_{ij}$, of the matrix $W$ is equal to the strength of directed association $i \to j$ if such association exists, and 0 otherwise (i.e. if the directed link $i \to j$ is absent).

Our main heuristic assumption is that to solve a RAT problem, a test subject performs a search on a free association network, which is a reflection of a search process happening in memory. Such a search process might imply, for example, exploration of all direct associations of all three stimuli words, or following chains of consequently extracted associations starting from stimuli words (such a chain may or may not be limited in length). More sophisticated search strategies can be used as well: for example, one may follow paths on network with preference of weak associations, or one may use some synergy between stimuli words (e.g. choosing words with strong associations with two or more stimuli words), etc. Finally, there exist a possibility that the solution is found but it is not recognized as the right one.

In order to test the basic hypothesis that the RAT solution is governed by some search process on the free association network, we study correlations between the RAT hardness and probabilities of finding a solution in various simple search strategies.

We begin with a simplest possible one-step strategy: (i) choose one of the stimuli words at random, (ii) jump to its neighbor along the directed link on the free association network (the jump probability is given by the link weight), (iii) check whether the solution is correct. The probability of finding a correct answer in such a strategy is, obviously,
\be
p_0 (\alpha)=\frac{1}{3}\left(w_{s_{\alpha}^1,r_{\alpha}}+w_{s_{\alpha}^2,r_{\alpha}}+ w_{s_{\alpha}^3,r_{\alpha}} \right)
\label{p_0}
\ee
where $\alpha$ enumerates different RAT problems ($1\le \alpha \le 138$), the indices $s_{\alpha}^1,s_{\alpha}^2,s_{\alpha}^3$ designate three stimuli words (vertices of the network) for a given problem $\alpha$, and the correct response is a network vertex with the index $r_{\alpha}$. Thus, $w_{s_{\alpha}^j,r_{\alpha}}$ is the weight of the directed bond $s_{\alpha}^j\to r_{\alpha}$, where $j=1,2,3$.

Another simple hypothetic model is as follows. Consider a search on the network as a sequence (Markov chain) of associations: one generates a random walk trajectory with jumping probabilities equal to $w_{ij}$, starting from one of stimuli words. In that case, the probability $\pi_{s,r}$ of reaching the response word from one of the stimuli is
\be
\pi_{s,r}= w_{s,r}+\sum_{k\neq r} w_{s,k} w_{k,r}+\sum_{k,l\neq r} w_{s,k} w_{k,l} w_{l,r} +\dots=\left[\frac{W^-}{I-W^-}\right]_{s,r},
\label{pi1}
\ee
where $W^-$ is the adjacency matrix $W$, in which all matrix elements $w_{r,i}$ are set to zero, which guarantees that only first passage of the response word is counted (the inverse elements, $w_{i,r}$, are kept in the matrix $W^-$). If the starting stimulus word is chosen at random, the resulting probability of solving the task by the proposed mechanism reads:
\be
p_1 (\alpha)=\frac{1}{3} \left(\pi_{s_{\alpha}^1,r_{\alpha}}+\pi_{s_{\alpha}^2,r_{\alpha}}+ \pi_{s_{\alpha}^3,r_{\alpha}} \right)
\label{p_1}
\ee

Every search is restricted in time. Therefore, the Markov chain representing the search on the network, should be finite. Thus, it seems reasonable to truncate the maximal length of search trajectories: if the search is not completed during the allowed time interval, we stop the search and start the new one from the same stimuli. Such a strategy resembles the random search with resetting \cite{majumdar_evans}. In case of a random resetting, the probability of solution in one search, given a stimulus $s$ and a response $r$, can be written as follows
\be
\pi_{s,r} (\lambda)=\left[\frac{W^-}{I-\lambda W^-}\right]_{s,r}, \qquad p_{\lambda} (\alpha)=\frac{1}{3}\sum_{i=1..3} \pi_{s_{\alpha}^i,r_{\alpha}}(\lambda)
\label{pi2}
\ee
with $\lambda$ being the resetting probability.

\begin{figure}[ht]
\centerline{\includegraphics[width=15cm]{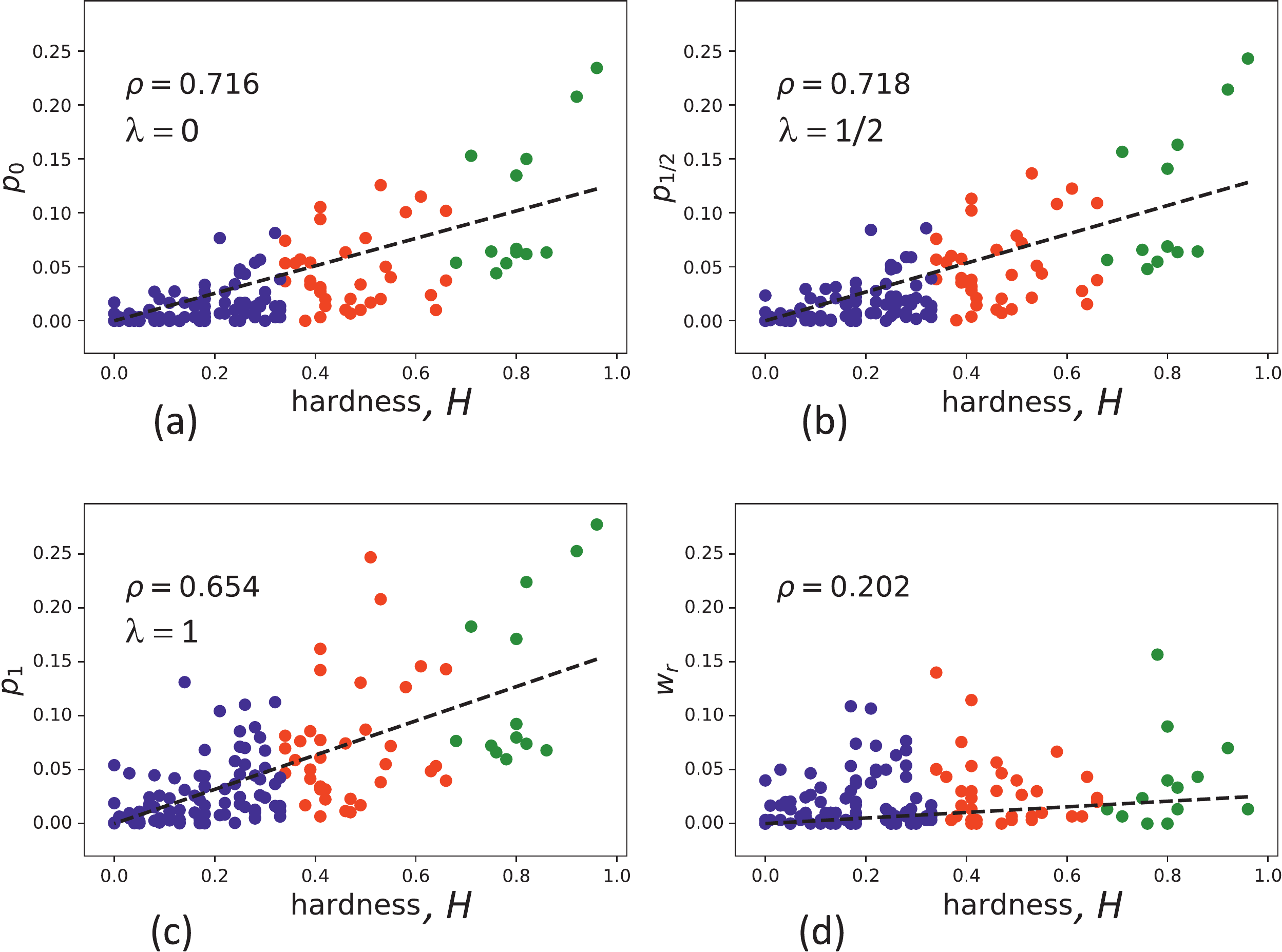}}
\caption{The scatter plots of the empirical hardness of the RAT problems versus following variables: (a) the average association weight from the stimuli words to the response $p_0(\alpha)$;  (b) the estimated probability of random walk with resetting with $\lambda={1/2}$, $p_{1/2}(\alpha)$; (c) the estimated probability of unlimited ($\lambda={1}$) random walk , $p_{1}(\alpha)$, (d) the average association weight from the response to the stimuli words $w_{\alpha}$. In all figures $\rho$ is the Pearson correlation coefficient.}
\label{fig02}
\end{figure}

In \fig{fig02}a-c scatter plots providing correlations between various search strategies and the empirical hardness, $H$, are shown. In particular, \fig{fig02}a presents the correlation between
the average association weight \eq{pi2} from stimuli words to the response, $p_{\lambda=0}$, and $H$; \fig{fig02}b -- the correlations between the estimated probability of the random walk with resetting, $p_{\lambda=1/2}$ and $H$; (c) -- the same as (b) for $p_{\lambda=1}$ and $H$. Dashed lines show slopes of the linear regression analysis, the corresponding Pearson correlation coefficient is designated by $\rho$. In all cases we observe sufficiently large values of $\rho$, which confirms our hypothesis that the RAT hardness correlates with relative locations of words in the associative network.

There is, meanwhile, another interesting question. The simplest strategies suggested above imply that solving RATs is a first-passage problem, that is to say, if the solver finds a solution, she immediately recognizes it. Is it indeed the case? One can address this question indirectly by studying correlation between the hardness of a RAT and the {\emph inverse} association weigths $w_{r_{\alpha},s_{\alpha}^i}$. Indeed, if the problem of recognizing an already found solution is at all relevant, it is natural to expect that it will be easier to recognize a solution for which there are strong associations from solution to the stimuli, and much harder to recognize ones for which there are no strong inverse associations. To check the existence of such a relation we plot in \fig{fig02}d the relation of the average inverse weight, $w_{\alpha}=\frac{1}{3} \sum_{i=1..3} w_{r_{\alpha},s_{\alpha}^i}$ and the RAT hardness. We see that the relation is very week (much weaker then the relation with direct weights), so in the first approximation one can assume that the problem of recognizing an already found solution is of secondary importance, as compared to the problem of finding the solution, and one can indeed treat solution of a RAT as a first passage problem.

Despite \fig{fig02} provides much important information, it does not reveal which search strategy is preferential. Indeed, $p_{\lambda}$ gives only a probability of finding a solution by a single Markov chain search, regardless its length. In reality, since the search is limited in time, the test subject might have enough time to try: either ten 1-step searches, or only one 10-step search. Moreover, one expects that there is a high variability how different test subjects solve problems. Thus, the question ``How to maximize the probability of solving a RAT?" seems to be more reasonable than the question ``How do people solve RATs on average?"

%We propose the hypothetic "optimal" search strategy in the following section, but before we proceed, one important question should be addressed. Is it indeed possible to consider the solution of a RAT as a \emph{first passage} problem? That is to say, if the test person finds the response word, does he/she immediately recognizes it as a solution? To check that, we have studied the correlation between the RAT hardness of a problem and \emph{reverse} weights $w_{r,s}$ along a path from the response to the stimulus. {\color{red} One expects that if recognizing a solution is not always easy, the solutions with high inverse rates (strong associations from the response to the stimuli) would turn out to be easier, while the solutions with small inverse rates would turn out to be harder. } In \fig{fig02}(d) we show the scatter plot of hardness of a RAT versus average inverse weight, $w_r=\frac{1}{3}\sum_{i=1..3} w_{r,s^i}$. The observed correlations are very weak, which confirms the hypothesis that the solutions are easily recognizable and the first passage concept is applicable.

\section{Enhancing the probability of correct solution}

Here we address the optimal strategy of maximizing the probability of solving a RAT problem. In particular, we are interested whether the heuristic optimal strategy depends on a RAT hardness. Clearly, two simplest strategies, \eq{p_0} and \eq{pi2}, outlined above, have significant drawbacks from that point of view. Searching only in the immediate proximity of a stimuli might by sufficient to solve easy RATs. But for hard RATs, there typically is no direct associations (direct links) between stimuli and response, thus solving a problem by such a strategy is simply impossible. In turn, searching via a random walk on a network might lead to excessively long solution times.

%Therefore, it seems natural to assume that one should restrict the search to one of the first "coordination spheres" of the stimuli, i.e. consider only the network vertices which are nearest neighbors,  next-to-nearest neighbors, next-to-next nearest neighbors, etc of a given stimulus. As an example, consider the following search process in the second coordination sphere.

Therefore, it seems natural to construct a search algorithm on a SWOW-EN network in a way that search trajectories, while not artificially constrained to nearest neighboring nodes of the stimuli, still are not fully random walks. One can think of these trajectories as of random walks in an external attractive potential, which guarantees that the test subject does not ever loose the stimuli words from his/her mental view. Such a strategy seems to be in agreement with the experimental data on sequences of guesses provided in \cite{smith2013} and discussed briefly in the Introduction.

\subsection{Search with an attraction to the stimuli}
\label{s:attr}

The proposed heuristic search algorithm on SWOW-EN is organized as follows. At time $t=0$ there are three stimuli (nodes of the network) $s^i$, $i=1,2,3$ which are considered active. At the next time step, $t=1$, one of nearest neighbors of the active nodes, $x$, is activated with probability $P(x)$ proportional to the sum of links from active network nodes towards it, i.e
\be
P(x)=\frac{\sum_{a}w_{a,x}}{\sum_{k}\sum_{a}w_{a,k}}, \quad (x,k) \in \mathbb{N}\mathbb{N}(\{a\}),
\label{activate}
\ee
where index $a$ enumerates active nodes, while index $k$ enumerates all possible target nodes from the set of nearest neighbors of the active ones $\mathbb{N}\mathbb{N}(\{a\})$.

Thus, at time $t=1$ there are four activated words. If the newly activated word is the correct response, $r$, the search is completed. If it is not, on the next step, $t=2$, we activate a new neighboring word with the probability \eq{activate} subject to the modification that now there are 4 instead of 3 active words in the set $\{a\}$. Simultaneously, we deactivate the word which was activated on a previous step, and mark it as checked, so that it will not be ever activated again. Now we check if the newly activated word is the correct response, $r$. If yes, we exit the search, if not, proceed recursively along with the procedure described above. At all times except $t=0$ there are exactly 4 active words, and by time $t$ exactly $t$ different possible response words have been checked.

By such rules we mimic a search strategy according to which the activated word, if it is not a response (i.e. a correct answer), still can affect the search trajectory leading to the answer. The fact that three stimuli remain always active at each time step, while intermediate guess words are activated and deactivated during a search, guarantees that there is an effective "attraction" of the search trajectory to the set of stimuli, which can be interpreted as a permanent "memory" about the initial stimuli.

\begin{figure}[ht]
\centerline{\includegraphics[width=15cm]{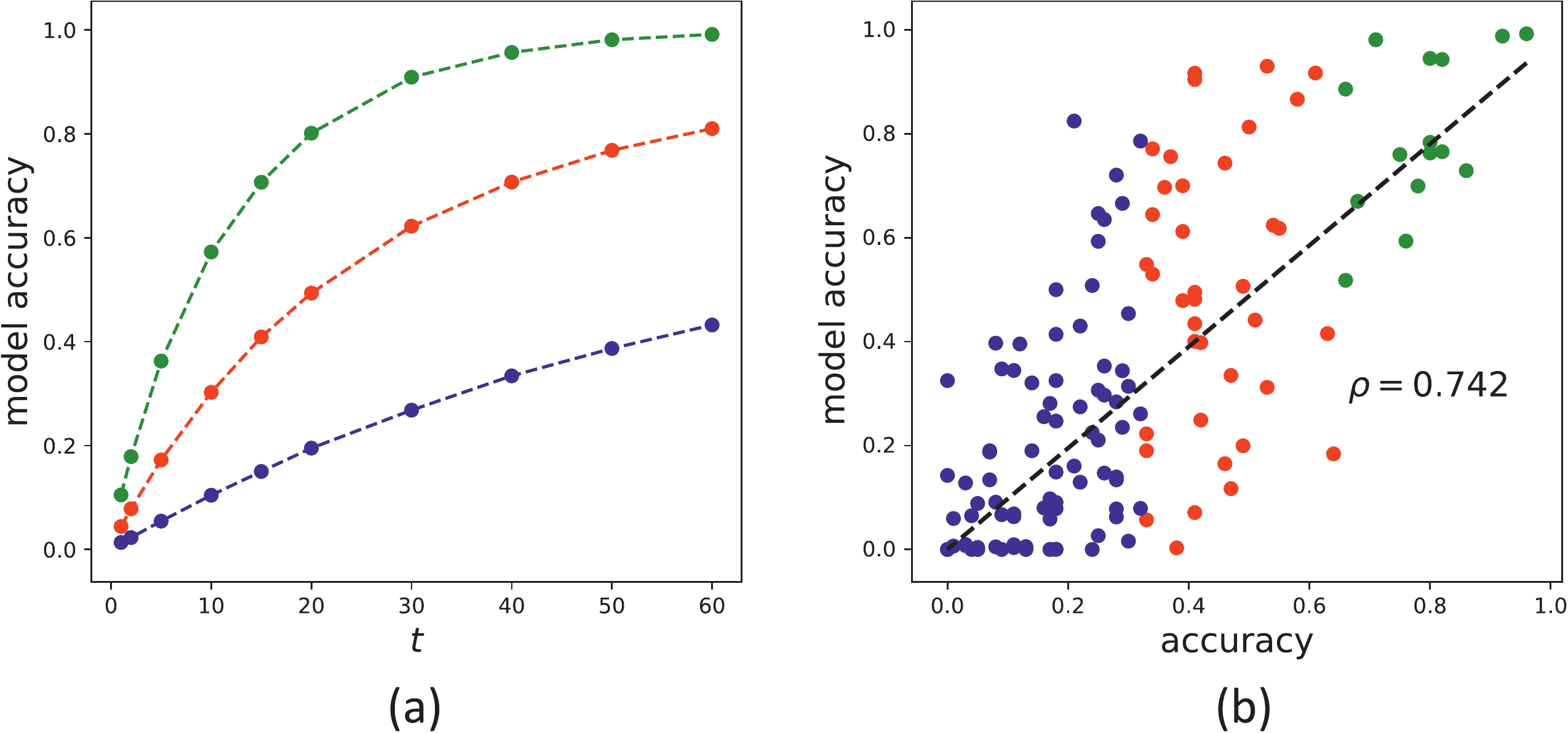}}
\caption{(a) The dependence of model accuracy on the number of search attempts (steps on SWOW-EN); (b) the scatter plot of model accuracy versus empirical hardness of the RAT. The model accuracy is averaged over $10^4$ simulations. The dashed line in (b) is the best linear fit to the data, $\rho$ is the corresponding Pearson correlation coefficient.}
\label{fig04}
\end{figure}

The search algorithm stops if either the correct answer is executed, or $t_{\max}$ search attempts are exceeded. We performed $10^4$ runs of that algorithm for each RAT, and counted the fraction of runs leading to the correct answer, this fraction is considered as the measure of the search accuracy. The resulting accuracy is a monotonously increasing function of $t_{\max}$ (see \fig{fig04}a) and at $t_{\max}=20$ the average accuracy of hard, medium and easy RATs numerically coincides with corresponding typical probabilities of correctly solved RATs in 15 seconds \cite{bowden2003}.

In \fig{fig04}b the scatter plot of the empirical RAT hardness versus the model search accuracy is shown. Note that the strategy presented in this section not only have more sense than those discussed in the previous one (the big majority of RATs are solvable by this procedure without relying on excesively long search trajectories), but also the correlation between experimental and model accuracy (the Pearson correlation coefficient, $\rho=0.742$) is larger than for those naive strategies. We believe therefore that this heuristic strategy is a better approximation to the way people solve RATs in reality.

\subsection{Activation algorithm with a threshold}
\label{s:activ}

Consider now a modification of our heuristic search algorithm described in Section \ref{s:attr}. It is known that many activation processes need a certain threshold (minimal activation impulse) to get trigged. In the psycholinguistic context, the importance of the association strength and the number of associations in the search processes is well known \cite{olteteanu2017}. In the spirit of \cite{olteteanu2017} we introduce an activation threshold to our model. We modify  \eq{activate} as follows
\be
P(x)= \left\{\begin{array}{cl}
\disp \frac{\sum_{a}w_{a,x}}{\sum_{k}\sum_{a}w_{a,k}} & \disp \mbox{if $\sum_{a}w_{a,x} \geq \tau, \quad (x,k) \in \mathbb{N}\mathbb{N}(\{a\})$} \medskip \\ 0 & \mbox{otherwise}
\end{array}\right. ,
\label{active_threshold}
\ee
Assume that a target $x$ can be activated only if the sum of activation signals exceeds a certain threshold, $\tau$.

\begin{figure}[ht]
\centerline{\includegraphics[width=15cm]{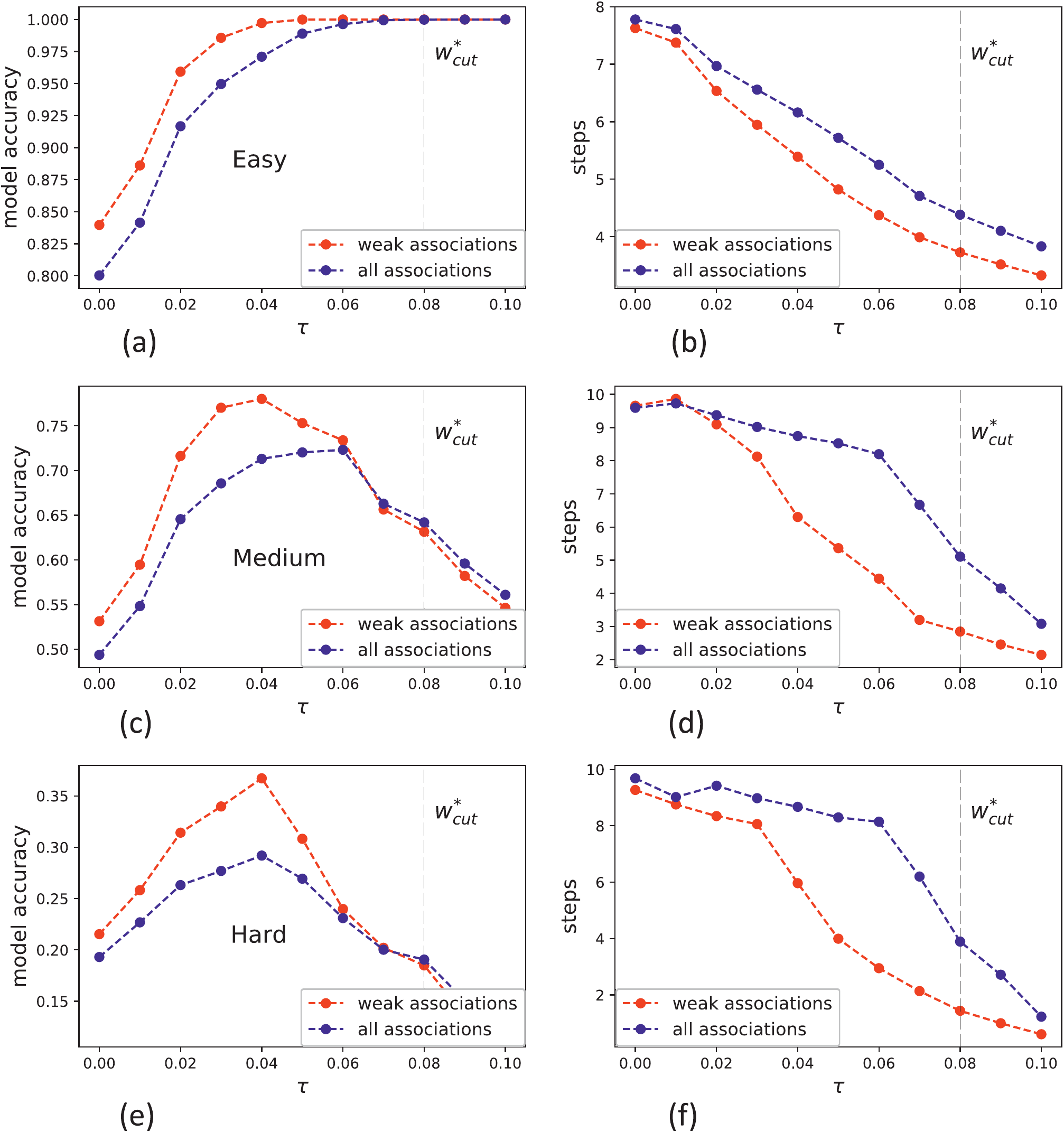}}
\caption{(a) The dependence of the predicted accuracy on the strength of the threshold $\tau$ for the RAT of different categories. The accuracy is calculated over $10^4$ simulations and averaged over all RATs in a given hardness category. (b) The dependence of the mean length of solving trajectories on the threshold strength.}
\label{fig05}
\end{figure}

We study the predicted accuracy for RATs of different hardness as a function of the threshold, $\tau$. For each $\tau$, the accuracy is averaged over $10^4$ simulations of all RATs in a given hardness category (easy/medium/hard). In \fig{fig05}a,c,e we show average predicted accuracy for different hardness categories as a function of $\tau$ ($\tau \in [0,0.1]$), while in \fig{fig05}b,d,f we show corresponding average times needed to solve easy/medium/hard RATs.

Solvability of easy RATs (\fig{fig05}a) grows monotonously and approaches unity with increasing $\tau$. It is easy to explain that: in easy RATs there is at least one strong directed link from a stimulus to the response, elimination of weak bonds makes such a strong link relatively stronger, which enhances the probability of a correct solution. The situation is different for medium and hard RATs. Eliminating very weak links (small $\tau$) leads to increased solvability similarly to easy RATs. That can be expected: presumably significant number of very weak bonds is just an experimental noise and its removal helps to reduce random and irrelavant trajectories. However, further increase of $\tau$ results in the accuracy of solution passing through a maximum at around $\tau_m=0.04$ (which is still below the percolation threshold corresponding to $w^*=0.08$ -- see \fig{fig01}c).

The probability of a correct solution at the maximum, $P(\tau_m)$, significantly exceeds the result of both a no-threshold model and of a model where only strong links are retained. Compared to the last model (with only strong links left), the gain is by a factor of 1.3 for medium RATs and by a factor of almost 2 for hard RATs. This means that moderately weak links are instrumental for solving medium and hard RATs: eliminating these moderate links decreases the solvability, and, as shown in \fig{fig05}b,d,f, increases the mean length of search trajectories.

\subsection{Enhancing the role of weak associations}

The result of Section \ref{s:activ} gives rise to the following natural question. Is it possible to enhance the solvability of medium/hard RATs further by {\emph preferentially} following moderately weak links? To check this, let us, apart from  removing weak links, remove also the strong ones. That is to say, introduce a new adjacency matrix $\bar{W}$ with matrix elements
\be
\bar{w}_{ij}= w_{ij} H(1-w_{\max}),
\ee
where $H(x)$ is the Heaviside function, and $w_{\max}$ is the upper cutoff parameter. Studying the behavior of the model \eq{active_threshold} for $w_{\max}=0.05$, we conclude that in that case the maximal accuracy for hard and medium RATs significantly increases (by a factor of 1.1 for medium RATs and by a factor of 1.3 for hard RATs) and the mean length of solving trajectories essentially decreases compared to the null model with $w_{\max}=1$ shown in \fig{fig05}c,e. The obtained result indicates the crucial role of moderately weak associations in the solution of medium and, especially, hard RATs.

\section{Discussion}

We have applied the network approach for studying the psycholinguistic mechanisms of solving Remote Association Tests (RATs). Our treatment is based on available open data on the network of free associations in English language (SWOW-EN) \cite{dedeyne2019}, and on standardized concept of hardness of RATs \cite{bowden2003}.

First, we have quantitatively characterized the correlation between the hardness of a particular RAT and the location of stimuli on the directed network of free associations. We have argued that hardness of a RAT is strongly correlated with the aggregated weight of bonds from the stimuli to the response, as well as with the aggregated weight of multistep chain of consecutive associations. On the other hand, we have not found any significant correlation between the RAT hardness and weights of reverse (response $\to$ stimuli) bonds.

Secondly, we have investigated the efficiency of RAT solutions using an activation algorithm which resembles the random walk in a potential well with attraction to the stimuli words of the RAT. We show that while for easy RATs the solution is mostly governed by strong associative bonds from stimuli to response, the solution of medium and especially hard RATs, is mostly determined by moderately weak bonds, i.e. bonds with weights about $w = 0.04 \pm 0.01$. Indeed, introducing an activation threshold we have seen that while neglecting very weak bonds is beneficial for the solution efficiency, neglecting moderately weak bonds suppresses the efficiency of finding the correct response. Even more, one can further enhance the solution probability for medium and hard RATs by removing strong bonds with weights larger than $w = 0.05$.

We see that "very weak" and "moderately weak" bonds behave differently in our consideration. That could be related to the fact the accuracy of measurement of very weak bonds in a free association network experiment is rather poor, and the significant number of very weak bonds is just experimental noise. So, the efficiency of the solution might be additionally increased by replacing the experimental free association network with a "cleaned-up" one in the spirit of \cite{peixoto18}. From a more general perspective, the importance of weak associations in the solution of RATs seems to be an example of the ubiquitous importance of weak ties in social sciences \cite{granovetter}.

To conclude, let us notice that there are numerous other standard tools of network treatment whose potential application to linguistic and psychological problems seems very promising. Spectral analysis is among the most effective modern approaches. Since the majority of semantic networks are directed, the eigenvalues of corresponding adjacency matrices are complex. The simplest objects attributed to the graph (network) spectrum are the spectral density and the level spacing distribution. Recently, the standard tools for the investigation of real spectra of symmetric adjacency matrices have been extended to complex spectra of non-symmetric matrices. Hence, we are hopefully well equipped to attack the spectral structure of RATs on directed networks. The corresponding spectral approach would provide the answers to questions associated with dynamic problems on directed networks, such as diffusion, localization and synchronization. The spectral analysis of a RAT problem will be presented in the forthcoming publication \cite{pospelov_inprep}.

\section*{Acknowledgements}

This work is supported by RFBR grant 18-29-03167. The authors are grateful to O. Morozova, A. Poddiakov and D. Ushakov for introducing us to the field of psycholinguistics, and to V. Avetisov, I. Kasyanov, E. Patrusheva, N. Pospelov, S. Tolostoukhova, and M. Zvereva for many interesting discussions on the structure and properties of various networks of free associations.

\begin{appendix}

\section{Hardness data for Remote Association Tests}

Here we provide the list of 138 RATs we have used in our research and their hardness according to the data of \cite{bowden2003}:

\begin{table}[ht]
\centerline{\includegraphics[width=15cm]{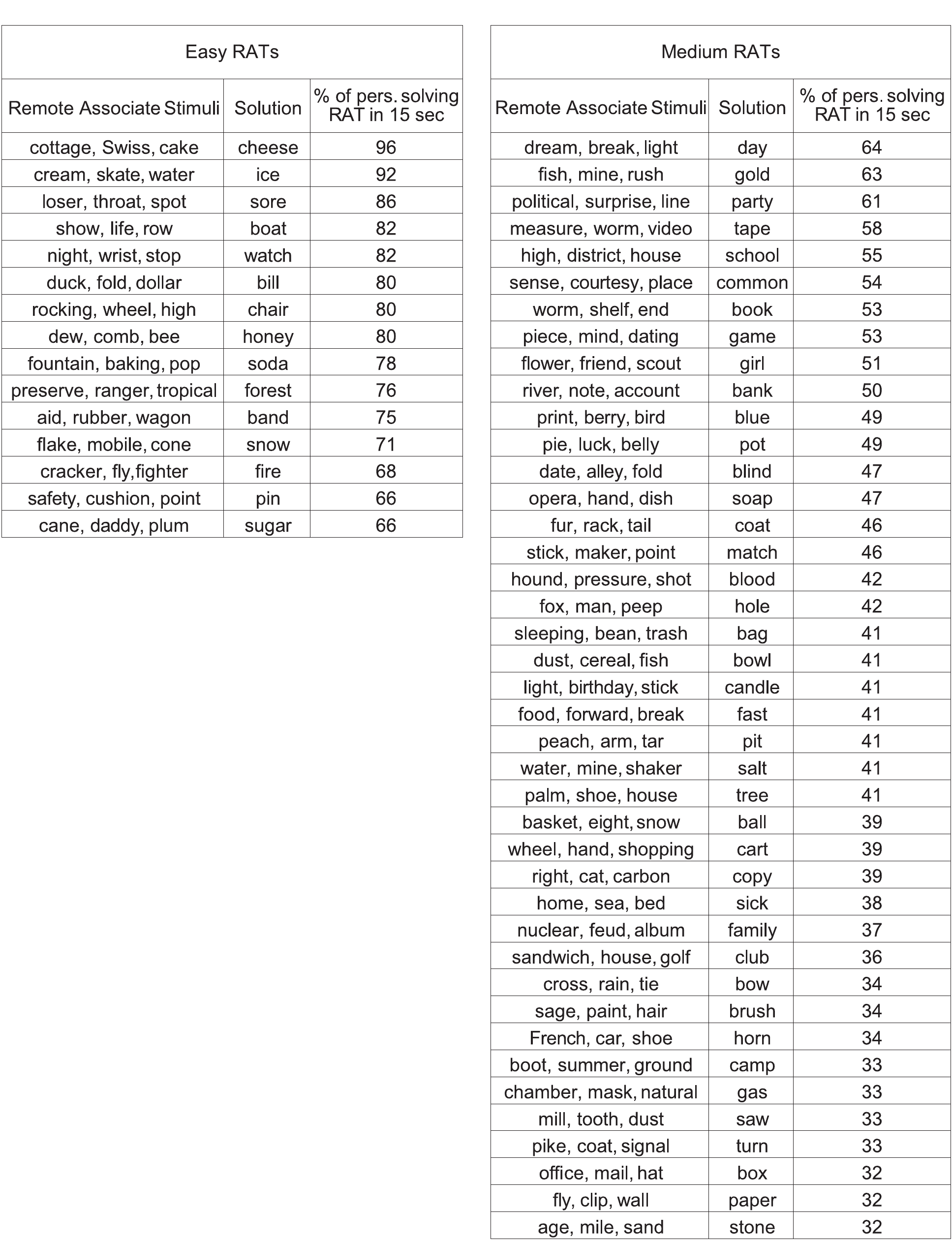}}
\label{tab01}
\end{table}

\begin{table}[ht]
\centerline{\includegraphics[width=15cm]{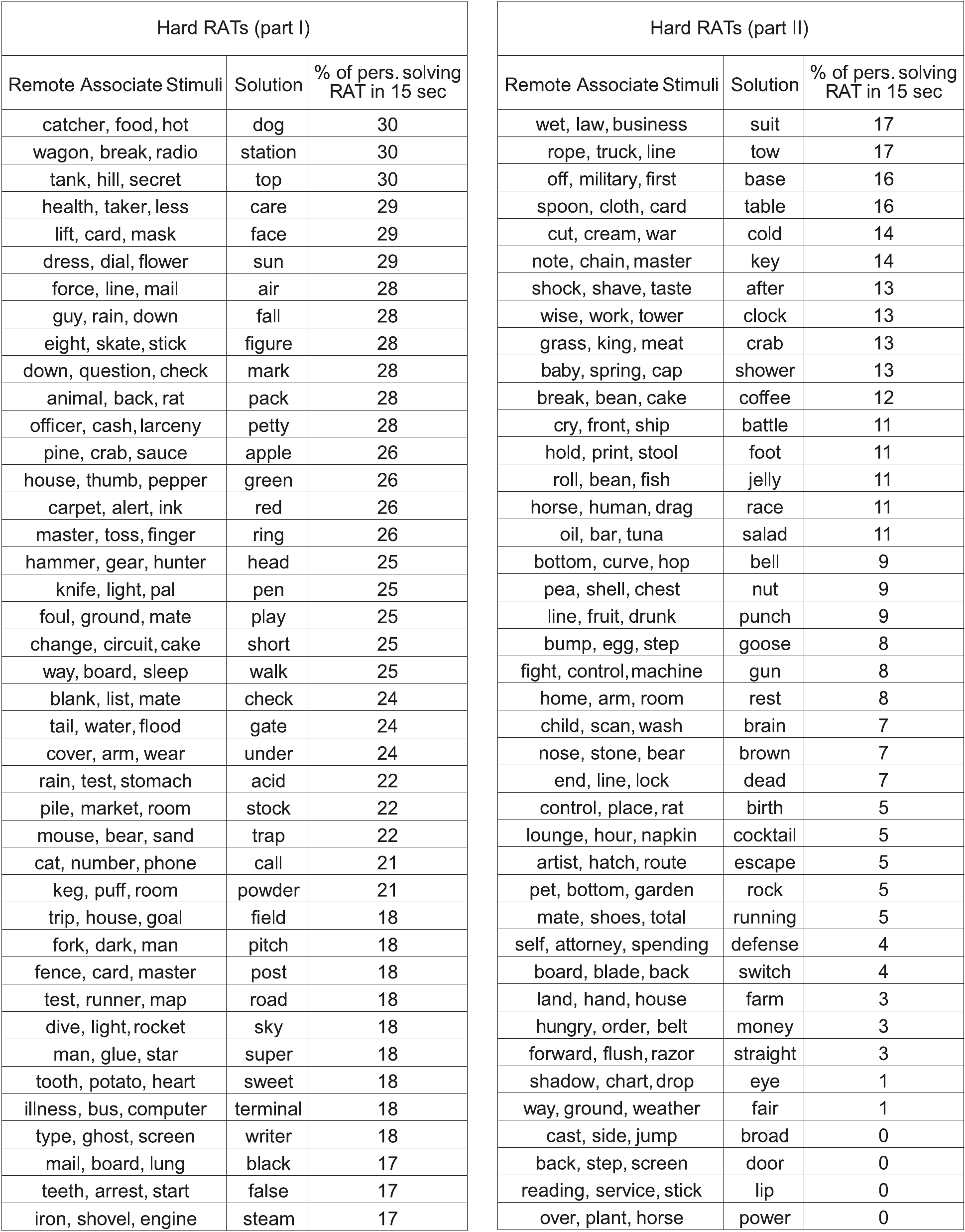}}
\label{tab02}
\end{table}

\end{appendix}

\end{document}